\documentclass[letterpaper, 10 pt, conference]{ieeeconf}  

\IEEEoverridecommandlockouts                              

\usepackage{graphics} 
\usepackage{epsfig} 
\usepackage{mathptmx} 
\usepackage{times} 
\usepackage{amsmath} 
\usepackage{amssymb}  
\usepackage{booktabs}

\usepackage[T1]{fontenc}
\usepackage{float}
\usepackage{subcaption}

\title{\LARGE \bf
Bi-VLA: Vision-Language-Action Model-Based System for Bimanual Robotic Dexterous Manipulations
}

\author{Koffivi Fidèle Gbagbe, Miguel Altamirano Cabrera, Ali Alabbas, Oussama Alyunes, \\ Artem Lykov,  and Dzmitry Tsetserukou%
    \thanks{The authors are with the Intelligent Space Robotics Laboratory, Center for Digital Engineering, Skolkovo Institute of Science and Technology, Moscow, Russia
    \tt \{Koffivi.Gbagbe, M.Altamirano, Ali.Alabbas, Oussama.Alyunes, Artem.Lykov, D.Tsetserukou\}@skoltech.ru}
}

\begin{document}

\maketitle
\thispagestyle{empty}
\pagestyle{empty}


\begin{abstract}

This research introduces the Bi-VLA (Vision-Language-Action) model, a novel system designed for bimanual robotic dexterous manipulation that seamlessly integrates vision for scene understanding, language comprehension for translating human instructions into executable code, and physical action generation. We evaluated the system's functionality through a series of household tasks, including the preparation of a desired salad upon human request. Bi-VLA demonstrates the ability to interpret complex human instructions, perceive and understand the visual context of ingredients, and execute precise bimanual actions to prepare the requested salad.
We assessed the system's performance in terms of accuracy, efficiency, and adaptability to different salad recipes and human preferences through a series of experiments. Our results show a 100\% success rate in generating the correct executable code by the Language Module, a 96.06\% success rate in detecting specific ingredients by the Vision Module, and an overall success rate of 83.4\% in correctly executing user-requested tasks.

\textbf{\small{\textit{Keywords--- Vision-Language-Action Model, Bimanual Robotic Manipulation, Human-Robot Interaction, Generative AI.}}}

\end{abstract}

\section{Introduction}
Recent advancements in language models have significantly impacted Human-Robot Interaction (HRI), enabling robots to engage more naturally and effectively with their human counterparts \cite{kim2024understanding}. In particular, to enhance performance efficiency and tackle complex tasks, coordinated dual-arm robotic systems are increasingly employed. These systems offer greater reliability and facilitate the execution of tasks that are challenging for a single manipulator \cite{vo2022development}. Furthermore, the coordination of two manipulator systems has been extensively studied in the literature, yielding satisfactory performance outcomes \cite{edsinger2007two}, \cite{rakita2019shared}.
However, traditional planning-based methods, which focus on motion planning through predefined trajectories for coordinating two robotic arms \cite{bersch2011bimanual}, \cite{mirrazavi2017dynamical}, are often suboptimal for dynamic and complex tasks. To achieve human-level bimanual manipulation, alternative strategies such as reinforcement learning (RL) \cite{chen2022towards} and learning from demonstration (LfD) \cite{franzese2023interactive} have been proposed. Nevertheless, it is crucial to acknowledge the inherent challenges associated with these approaches, particularly regarding the time required for learning and data collection. 
 \begin{figure}[t!]
    \centering
    \includegraphics[scale=0.64]{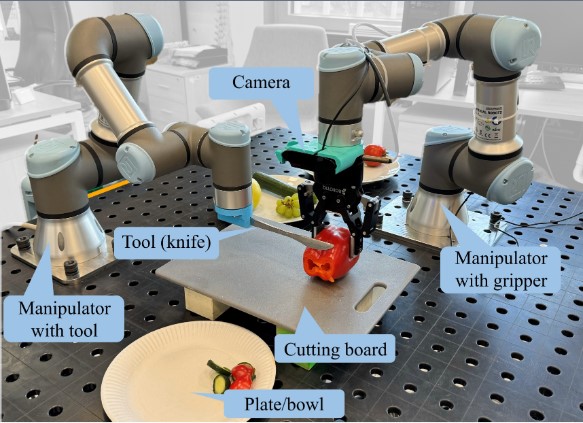}
    \caption{System Overview of Bi-VLA.}
    \label{system_overview}
\end{figure}

Moreover, the integration of language models in HRI not only facilitates natural communication but also extends their utility beyond basic interactions. This includes tasks such as task planning and generating intermediate reasoning steps, commonly referred to as a chain of thought. However, despite their potential, the application of language models to synthesize the bimanual skills of robots has not received significant attention. This is particularly noteworthy given that numerous tasks humans perform routinely require the coordinated use of both hands, highlighting the essential role of bimanual capabilities in executing intricate operations that are beyond the reach of single-arm manipulators.
Building on this premise, the integration of vision, language, and action within a unified model presents a novel approach to enhancing robotic bimanual operation. In this research, we aim to develop a vision-language-action model, dubbed Bi-VLA, specifically designed for bimanual robotic operations. Additionally, we will explore its potential impact on coordinating the actions of dual-arm robotic systems in response to human instructions. The proposed scenario is a cooking robot, where a robot is in charge of picking and placing the required ingredients, and the second cutting them. 


\begin{figure*}[!ht]
    \includegraphics[scale=0.055]{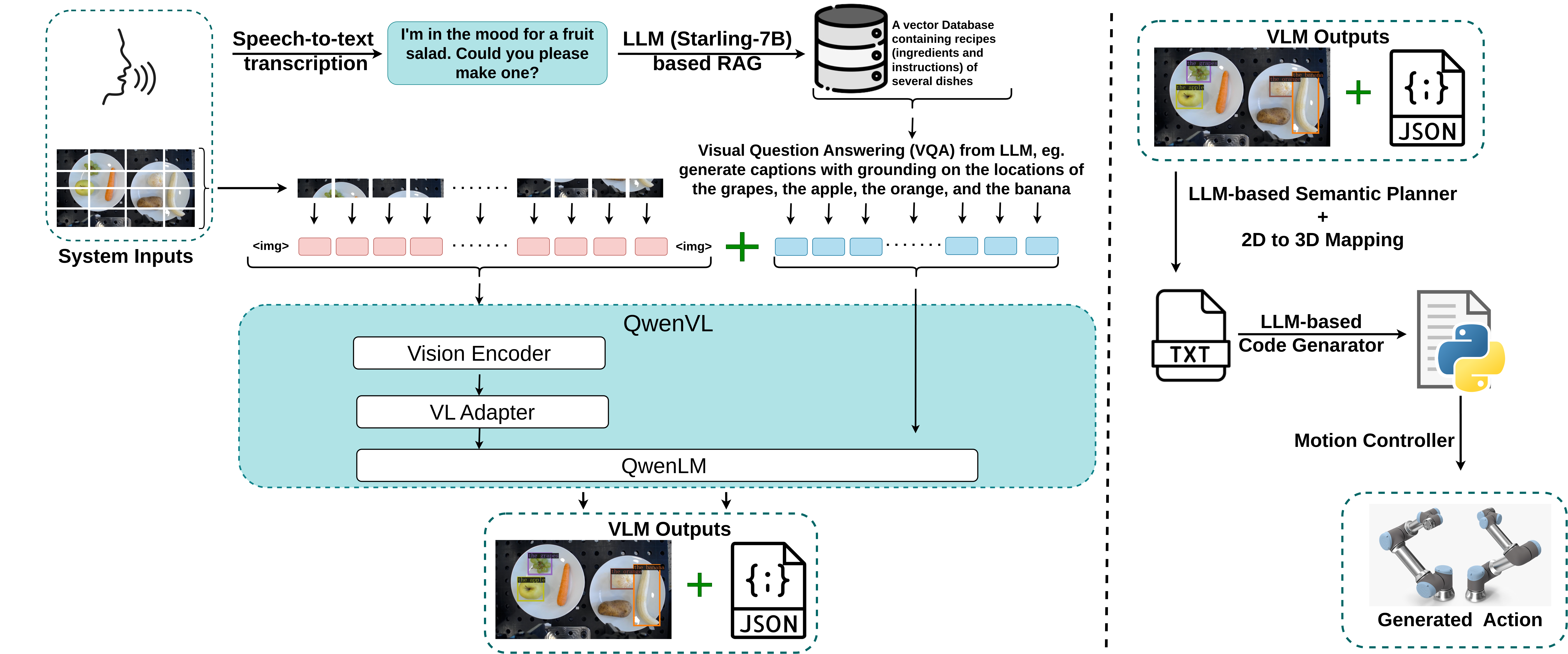}
    \caption{Bi-VLA Architecture.}
    \label{architecture}
\end{figure*}

\section{Related Work}
\subsection{From language instructions to robotic actions}
Translating user language input into robotic actions has seen considerable progress through the integration of Large Language Models (LLMs) and various algorithmic frameworks. CLIPort \cite{shridhar2022cliport} leverages visual and language models for robotic manipulation, focusing on the spatial understanding necessary for accomplishing the tasks. Similarly, ProgPrompt \cite{singh2023progprompt} introduced a programmatic LLM prompt structure that enables plan generation for robotic tasks scoring potential actions without the need for explicit domain knowledge. The ``Code as Policies" \cite{liang2023code} paradigm further underscores the importance of programmatic structures in embodying LLM-generated instructions for control, which is in close alignment with RT-1's objective of scalable, real-world robotic control \cite{brohan2022rt}. 
Drawing on pre-trained vision-language models, open-world object manipulation \cite{stone2023open} is evolving to adapt robotic actions to diverse and changing environments. Instruct2Act \cite{huang2023instruct2act} steps beyond basic instructions by aligning multimodal inputs with actionable robotic sequences, aided by the breadth of responses offered by LLMs. ``Language to Rewards" \cite{yu2023language} introduced a method of reward function generation from natural language inputs to calibrate and optimize robotic skills, with the goal of building cooperative agents by harnessing modularity through LLM architectures.
Continuing this thread, RT-2 \cite{rt22023arxiv} focuses on translating visual perceptions and user language into actionable instructions, demonstrating the critical intersection between understanding and physical task execution. 
Lastly, PaLM-E\cite{driess2023palm} highlights the convergence of multimodal understanding within LLMs, emphasizing the embodiment of language models in tangible robotic actions, while LLM-MARS\cite{lykov2023llmmars} incorporates LLMs into the multi-agent robotics system for behavior trees generation. Collectively, these advancements contribute to an overarching framework that not only enhances robots' ability to understand and act on complex user language inputs but also redefines interactions between humans and robots.
In this work, we use LLMs as a strategic semantic planner, orchestrating the collaborative efforts of robotic arms to successfully complete the given task.

\subsection{Bimanual robotic manipulation}
The exploration of bimanual object manipulation spans a multidisciplinary field, incorporating insights from neuroscience, robotics, and computer science. To establish a foundation, researchers have proposed various taxonomies aimed at emulating human-like coordination in complex tasks. Notably, the integration of spatial and temporal constraints is essential for achieving dexterity \cite{krebs2022bimanual}. 
Moreover, behavioral control paradigms suggest that dual-arm coordination often surpasses single-arm methods, highlighting the necessity for behavior-based control systems to facilitate effective bimanual interaction, as demonstrated by studies utilizing deep learning for imitation learning that illustrate effective teaching methods for robots through the observation and replication of human actions \cite{xie2020deep}.
Additionally, low-cost teleoperation solutions, such as Mobile ALOHA \cite{fu2024mobile}, have opened avenues for learning complex manipulation behaviors. Finally, progress in this area is complemented by advancements in sim-to-real deep reinforcement learning, which refines bimanual tactile manipulation abilities in simulated environments before transferring them to real-world applications \cite{lin2023bi}. 
In this work, we focus on a less data-intensive approach that eliminates the need for training or additional fine-tuning. This method involves the language module invoking specific predefined action APIs, determining their calling sequence, and allocating tasks between the two robotic arms to complete the user’s assigned task.

\section{Grounding Language Instructions into Bimanual Robot Actions}
\subsection{General system overview}
The system takes as input the user’s task request, which is transcribed into a text message, along with an image from the camera (see Fig. \ref{architecture} left). The task request serves as a query for the Retrieval-Augmented Generation (RAG) system, and its output initiates a visual question-answering (VQA) dialogue between the Language Module and the Vision Language Module. The input image, on the other hand, is split into patches, tokenized, and combined with the VQA question tokens before being sent to the Vision Language Module. The main outputs include an image with captions and a JSON file containing the 2D positions of ingredients, a list of all ingredients, and missing ingredients, etc. 
Subsequently, the Language Module generates a set of actions, which are performed by the robots (see Fig. \ref{architecture} right). In Fig. \ref{experiment} an example of a set of actions can be observed when all the required ingredients are available. In Fig. \ref{semantic_planning} the details of a single action are shown.

\subsection{Large Language Model Module}
In this work, we utilized \textbf{Starling-LM-7B-alpha} \cite{starling2023}, an open-source LLM developed by UC Berkeley's NEST Lab using Reinforcement Learning from AI Feedback (RLAIF), as the backbone for the language module, which outperforms \textbf{GPT-3.5} \cite{lib:openai2022introducing} in several benchmarks, including an 8.09 out of 10 in MT Bench.

\subsubsection{Semantic planning}\label{Semantic__planning}
Semantic planning involves the use of natural language understanding and generation techniques to interpret human instructions and generate executable plans. In the context of our work, the semantic planner is designed to generate a plan for coordinating the movements of the two robot arm manipulators equipped with different and appropriate tools in order to perform a given action. The planner takes the action description as input and produces a plan specifying the sequence of motions that need to be executed by the robot arms (see Fig. \ref{semantic_planning}). The planner utilizes a template that provides a general structure for the plan. It starts with the ``[start of plan]" tag and ends with the ``[end of plan]" tag, ensuring that the plan is clearly delineated. The template also includes specific instructions and placeholders that need to be filled based on the task requirements by the vision-language model (VLM) (see section \ref{qwen}), as well as a set of rules to follow.


\begin{figure*}[!ht]
    \includegraphics[scale=0.77]{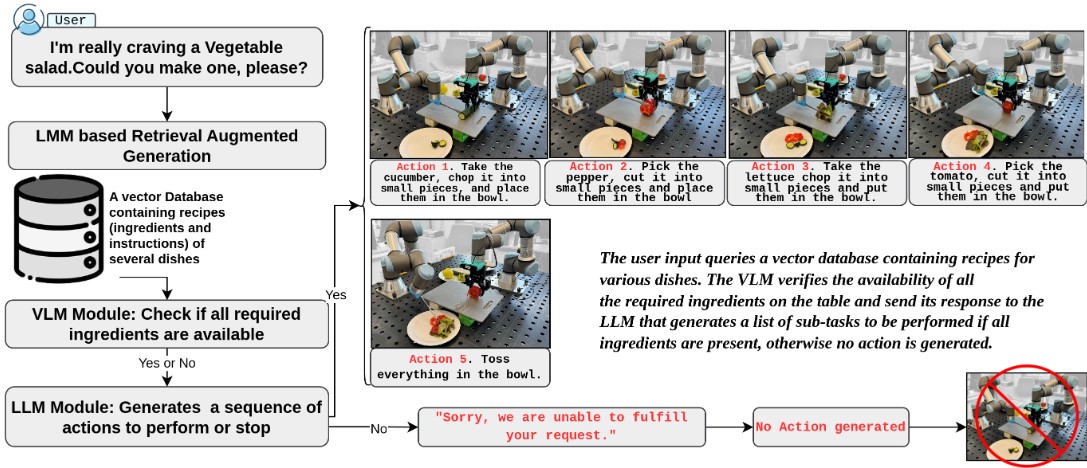}
    \caption{Cooking Experiment Workflow.}
    \label{experiment}
\end{figure*}


\subsubsection{Translating semantic plan into APIs function calls with the Code Generator}
\begin{figure*}[!ht]
    \includegraphics[scale=0.1150]{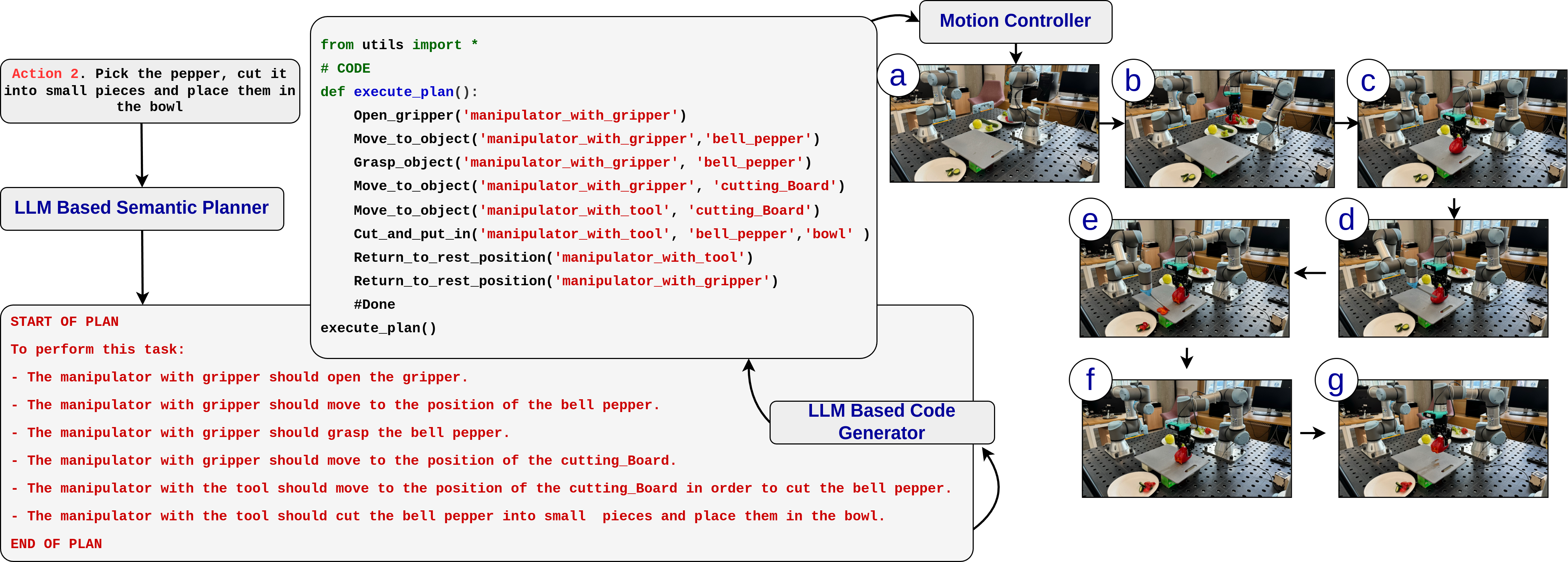}
    \caption{ Example illustration of Action 2 in Fig. \ref{experiment}. The LLM-based semantic planner receives the action to perform and generates a detailed plan outlining the movements of two robot manipulators. The generated plan is translated into a set of motion API call functions through the LLM-based Code Generator. The execution of the generated code triggers the movement of the two robot manipulators through the Motion Controller.(a)-(b) the manipulator with the gripper moves to grasp the pepper and bring it to the cutting board; (c)-(d) the manipulator with the knife moves to the cutting board, cuts the pepper (e), and places it in the bowl; (f)-(g) the two manipulators return to their initial position.}
    \label{semantic_planning}
\end{figure*}

The plan generated by the semantic planner is translated into a sequence of API call functions. The goal is to convert the textual plan into executable Python code that adheres to a predefined set of functions provided by the API. We use a prompt that outlines the functions available for use, the structure of the plan, and the rules to follow. To illustrate the process, we provide an example plan and its corresponding Python code (Fig. \ref{semantic_planning}), helping to guide the model's response as in the one-shot prompting technique. This methodology facilitates the seamless translation of high-level task plans into executable Python code, promoting efficiency and clarity.


\subsubsection{From API function call to Action Generation with The Motion Control APIs}
To facilitate action generation, we have implemented a comprehensive set of API functions to control the robot's actions. These functions are designed to perform a variety of tasks, including:
\begin{itemize}
    \item \textbf{open\_gripper(manipulator\_type)}: to open the gripper.
    \item \textbf{move\_to\_object(manipulator\_type, object\_name)}: to move the manipulator to the position of an object.
    \item \textbf{grasp(manipulator\_type, object\_name)}: to grasp an object.
\end{itemize}
\begin{itemize}
    \item \textbf{cut(manipulator\_type, object\_name)}: for cutting actions.
    \item \textbf{pour(manipulator\_type, object\_name)}: for pouring actions.
    \item \textbf{put(manipulator\_type, object\_name)}: for pick-and-place actions.
    \item \textbf{toss(manipulator\_type, object\_name)}: for mixing objects or vegetables in a bowl.
    \item \textbf{cut\_and\_put\_in(manipulator\_type, object\_name)}: which combines the \textbf{cut} and \textbf{put} actions.
\end{itemize}
The vision language model uses the function \textbf{get\_list\_of\_objects()} to retrieve the names of all objects visible on the table within the camera frame, as well as \textbf{get\_bounding\_boxes()} which generates the bounding boxes of the desired list of objects. Ensuring explicit naming of API functions is crucial to preventing confusion with the language model. This comprehensive set of API functions provides the necessary control and versatility for robotic operations, enabling a wide range of complex tasks to be performed with precision and efficiency.

\subsection{Vision Language Model Module}
\subsubsection{Integration of the Qwen Vision-Language Model}
\label{qwen}
The QWEN-VL series are large-scale vision-language models (VLMs) that understand both text and images \cite{Qwen-VL}. Built on the QWEN-LM foundation, they feature visual capabilities through a specialized architecture and training pipeline, achieving state-of-the-art performance across various benchmarks, including image captioning, visual question answering, and visual grounding. In this work, we use the model to verify the availability of required items and provide the 2D-pixel coordinates of these items.

\subsubsection{Mapping Image Pixel to 3D Coordinates} \label{2d}
We assumed that the position of an object is equal to the position of its center. The relationship between the position of an object in the world coordinate system and pixel coordinates is given by the equation:
\begin{equation}
    z_c \ ^pP=^p_cT \ ^c_w\Grave{T} \ ^wP,
    \label{eq:world2pixel_coordinates}
\end{equation}
where $z_c$ is the position of the object on the $z$ axis of the camera coordinates; $^pP = [x_p \ y_p \ 1]^T$ is the pixel coordinate of the object; $^p_cT$ is the intrinsic matrix of the camera, a $3 \times 3$ matrix, that transforms position vectors from the camera coordinates to the pixel coordinates, and we get it when we calibrate the camera; $^c_w\Grave{T}$ is the $3 \times 4$ matrix that transforms position vectors from world coordinates to camera coordinates, and it is known since the camera is fixed on the robot tool and the transformation between the tool and the world coordinates is known all the time, and $^wP = [x_w \ y_w \ z_w \ 1]^T$ is the position of the object in the world coordinates.

We rewrite $^c_w\Grave{T}$ as:
\begin{equation}
^c_w\Grave{T} = \left[
\begin{array}{c|c}
^c_wR_{3 \times 3} & ^c_wt_{3 \times 1}
\end{array}
\right],
\end{equation}
where $^c_wR$ is the rotation matrix from the world coordinates to the camera coordinates and $^c_wt$ is the translation vector. We also define 
\begin{equation}
^c_wT = 
\left[
\begin{array}{cccc}
 & ^c_w\Grave{T} & &\\
\midrule
0 & 0 & 0 & 1
\end{array}
\right] = 
\left[
\begin{array}{c|c}
^c_wR_{3 \times 3} & ^c_wt_{3 \times 1}\\
\midrule
0_{1 \times 3} & 1
\end{array}
\right].
\end{equation}

The pixel position that we get from the camera is distorted with radial and tangential errors \cite{de2008centi}. In order to obtain accurate transformation from pixel to real-world coordinates, these distortion errors have to be taken into account. When the camera was calibrated, we obtained the intrinsic matrix $^p_cT$ and the distortion coefficients $[k_1, k_2, p_1, p_2, k_3]$. To correct the distortion error, we used the Brown–Conrady model that follows the following equations \cite{conrady1919decentred}:

\begin{equation}
\begin{aligned}
x_u = & x_d(1 + k_1 r^2 + k_2 r^4 + k_3 r^6) + \\
      & + 2p_1x_dy_d + p_2(r^2+2x_d^2), \\
y_u = & y_d(1 + k_1 r^2 + k_2 r^4 + k_3 r^6) + \\
      & + p_1(r^2+2y_d^2) + 2p_1x_dy_d,    
\end{aligned}
\label{eq:undistorted_pixels}
\end{equation}
where $x_u$ and $y_u$ are the undistorted pixel coordinates, $x_d$ and $y_d$ are the distorted pixel coordinates, $[k_1, k_2, p_1, p_2, k_3]$ are the distortion coefficients and $r = \sqrt{x_d^2+y_d^2}$ is the Euclidean distance of the distorted point.

To transfer the position of the objects from Pixel to 3D coordinates, we followed the following steps:

\begin{enumerate}
    \item Get the distorted pixel position of the center of the object $[x_d \ y_d]^T$.
    \item Calculate the undistorted pixel coordinates $[x_u \ y_u]^T$ from the distorted ones using Eq. \ref{eq:undistorted_pixels}.
    \item Calculate the normalized coordinates in the camera coordinates system depending on Eq. \ref{eq:world2pixel_coordinates}.

    \begin{equation*}
        \begin{bmatrix}
           \frac{x_c}{z_c} \\
           \frac{y_c}{z_c} \\
           1
        \end{bmatrix} = ^p_cT^{-1}        
        \begin{bmatrix}
            x_u \\
            y_u \\
            1
        \end{bmatrix},
    \end{equation*}
    where $x_c, y_c, z_c$ are the position of the object center in the camera coordinates system.
    
    \item Calculate the real position in the camera coordinates, $[x_c \ y_c \ z_c]^T$, from the normalized vector by multiplying it by $z_c$. In our application all objects are placed on the table in the same $z-plane$ relative to the camera coordinates system and the camera is operating in a known altitude from the table and its $z$ axis is perpendicular to the table.

    \item Calculate the position of the object in the world coordinates $[x_w \ y_w \ z_w]$ from the position in the camera coordinates.
    \begin{equation*}
        \begin{bmatrix}
            x_w \\
            y_w \\
            z_w \\
            1
        \end{bmatrix} = ^c_wT^{-1}        
        \begin{bmatrix}
            x_c \\
            y_c \\
            z_c \\
            1
        \end{bmatrix}
    \end{equation*}
\end{enumerate}

These steps allow us to get the world coordinates of the objects, which are then sent to the robot. After receiving the position, the robot moves its Tool Center Point (TCP) above the object and then gets it down to catch the object.



\section{Experimental Evaluation}
\subsection{Experimental Setup}  
Two UR3 collaborative robots from Universal Robots were mounted on a Siegmund welding table (see Fig. \ref{system_overview}). A custom-designed and 3D-printed camera holder, made of PLA, was attached to the first manipulator, which also held a Logitech HD 1080p webcam and a two-finger Robotiq gripper. A knife holder was similarly designed and installed in the second manipulator's end effector. Ingredients for various salads were positioned near the first manipulator where the camera was able to capture images to identify the necessary items. A cutting board was placed between the two manipulators for cutting operations, with a plate or bowl in front to place the prepared ingredients. All processes are supervised by a PC 
(NVIDIA GeForce RTX 4090 24Gb, SSD 2Tb, RAM 64Gb 5600MHz, Intel Core i9, running Ubuntu 22.04).


\subsection{Experiments Workflow}
As depicted in Fig. \ref{experiment}, we design several custom cooking recipes stored in a vector database, which serves as a set of tasks for the robot manipulators to perform collaboratively. The recipes clearly state the required ingredients as well as the instructions to follow for successful execution. Among these sets of tasks, we evaluate the performance of our system on three selected tasks, including making a vegetable salad, a Russian salad, and a fruit salad. The three tasks are different in nature in the sense that they require several types of actions, such as picking, placing, cutting, pouring, mixing, etc. 
When a user submits a request, it serves as a query processed through the LLM-Based Retrieval-Augmented Generation (RAG) technique \cite{lewis2021retrievalaugmented}, which enhances response quality and factuality by augmenting the language model's outputs with relevant information from external knowledge sources , in this case, a vector database.
After the RAG model produces a response, the VLM  module verifies the availability of all the ingredients mentioned in the recipe. Based on the VLM's response, the LLM module will then generate a list of actions for the two robots to perform if the ingredients are available. Alternatively, if the ingredients are not available or the request is ambiguous, no action will be performed.
\section{Results and Evaluation}
We assessed the performance of our system's components both individually and collectively, enabling us to identify their strengths and weaknesses and understand their interactions in contributing to overall performance. We define a task as completed successfully if the generated code by the Code Generator is correct before execution by the Motion Controller and if the actions are successfully executed.

\begin{enumerate}
    \item  

Firstly, we evaluated the performance of the Language Module by generating 100 different requests with GPT-3.5 for the system to prepare three types of salad, e.g. ``I'm really craving a vegetable salad. Could you make one, please?". Specifically, one-third of the requests were for making vegetable salads, another third were for a Russian salad, and the final third was for a fruit salad. Our proposed Language Module successfully generated the correct executable code for all the requests, achieving a success rate of 100\%.

\item  
In the second part of the experiment, we evaluated the vision module's performance using 100 pictures of different ingredients, with one-third allocated to each salad type.
For each salad type, half the pictures included all the required ingredients (Scenario 1), while the other half were missing some (Scenario 2). We evaluated if the model returned the correct list of all the ingredients in the input image and whether each mentioned ingredient from the salad recipe was detected correctly; that is, the captions on the ingredients and the 2D bounding boxes were accurate in the output image.

\begin{itemize}
    \item In Scenario 1, where all ingredients were correct, we achieved a 96.06\% success rate in detecting ingredients and an 83.4\% success rate in generating correct actions.
    
    \item In Scenario 2, the vision module misidentified present ingredients as missing ones, leading to a detection performance of 86.53\% and a 71.22\% success rate in generating correct actions.
\end{itemize}

\item For the Action Generation Module, we ensured reliable translation from 2D pixel to 3D world coordinates as described in Section \ref{2d}, enabling the robot's gripper to successfully grasp objects of varying orientations and perform cutting operations accurately. Therefore, we assume that the success rate of the module is equivalent to the success rate of generating correct 2D bounding box coordinates. 
\end{enumerate}

The table below summarizes  the main experiment results:






\begin{table}[!htb]
\centering
\caption{Experiment Results}
\begin{tabular}{cccc}
    \toprule
    Experiments      & Success rate (\%)\\
    \midrule
    LLM Module (all scenario)                   &  \quad\quad  100\% \\
    VLM Module (scenario 1)    & \quad\quad 96.06\%\\
    VLM Module (scenario 2)   & \quad\quad 86.53\% \\
    Action Generation Module (scenario 1 )  & \quad\quad 83.4\% \\
    Action Generation Module (scenario 2) & \quad\quad 71.22\%  \\
    \bottomrule
    \end{tabular}
    \label{tab:2}
\end{table}



\section{Conclusion}
In this work, we explore a new approach to dexterous bimanual manipulation by integrating vision, language understanding, and robotic action generation into a unified system named Bi-VLA. Our experiments demonstrate the system's capability to accurately interpret human instructions, perceive the visual context of the scene, and execute the required actions. The high success rates of the language and vision modules emphasize their crucial roles in overall system performance. Our findings highlight the need for ongoing advancements in visual understanding, as the accuracy of the vision module significantly affects task completion. Future efforts will focus on enhancing the robustness and versatility of the vision module to better handle a variety of visual variations and uncertainties, ultimately making robotic assistants more reliable and adaptable.

\section*{Acknowledgements} 
Research reported in this publication was financially supported by the RSF grant No. 24-41-02039.

\bibliographystyle{IEEEtran}
\bibliography{mybib}

\end{document}